\pdfoutput=1

\documentclass[11pt]{article}

\usepackage[final]{acl}

\usepackage{times}
\usepackage{latexsym}

\usepackage[T1]{fontenc}

\usepackage[utf8]{inputenc}

\usepackage{microtype}

\usepackage{inconsolata}

\usepackage{graphicx}

\title{IITR-CIOL@NLU of Devanagari Script Languages 2025: Multilingual Hate Speech Detection and Target Identification in Devanagari-Scripted Languages}

\author{
Siddhant Gupta\textsuperscript{1}, Siddh Singhal\textsuperscript{1}, and Azmine Toushik Wasi\textsuperscript{2}\\
\textsuperscript{1}Indian Institute of Technology, Roorkee, India\\
\textsuperscript{2}Shahjalal University of Science and Technology, Sylhet, Bangladesh \\
  \texttt{siddhant\_g@me.iitr.ac.in, siddh\_s@ee.iitr.ac.in, azmine32@student.sust.edu} \\
  \small{All authors contributed to the work equally.}}

\begin{document}
\maketitle
\begin{abstract}
This work focuses on two subtasks related to hate speech detection and target identification in Devanagari-scripted languages, specifically Hindi, Marathi, Nepali, Bhojpuri, and Sanskrit. Subtask B involves detecting hate speech in online text, while Subtask C requires identifying the specific targets of hate speech, such as individuals, organizations, or communities. We propose the \texttt{MultilingualRobertaClass} model, a deep neural network built on the pretrained multilingual transformer model \texttt{ia-multilingual-transliterated-roberta}, optimized for classification tasks in multilingual and transliterated contexts. The model leverages contextualized embeddings to handle linguistic diversity, with a classifier head for binary classification. We received 88.40\% accuracy in Subtask B and 66.11\% accuracy in Subtask C, in the test set.
\end{abstract}

\section{Introduction}
South Asia—comprising Afghanistan, Bangladesh, Bhutan, India, Maldives, Nepal, Pakistan, and Sri Lanka—is one of the world’s most populous regions, with around 1.97 billion people \cite{alexeeva2024southasia}. This region is incredibly rich in culture and language, with over 700 languages and about 25 major scripts in use. More than 50 million South Asians also live abroad, further spreading this linguistic diversity \cite{Rajan2007}. 

Despite this, South Asian languages are underrepresented in language technology. Many recent large language models (LLMs) have very limited data from South Asia, which is partly due to the numerous challenges in processing these languages \cite{nguyen-etal-2024-seallms}. Encoding issues are an early hurdle; although most South Asian scripts are Unicode-compliant, many applications struggle to render them correctly due to complex orthographies, and language input remains a challenge. The region’s languages also have intricate linguistic features, multiple writing systems, and long-standing literary traditions that make natural language processing (NLP) difficult \cite{joshi2022l3cubehind}.

\begin{figure}[t] 
\centering {\includegraphics[scale=0.18]{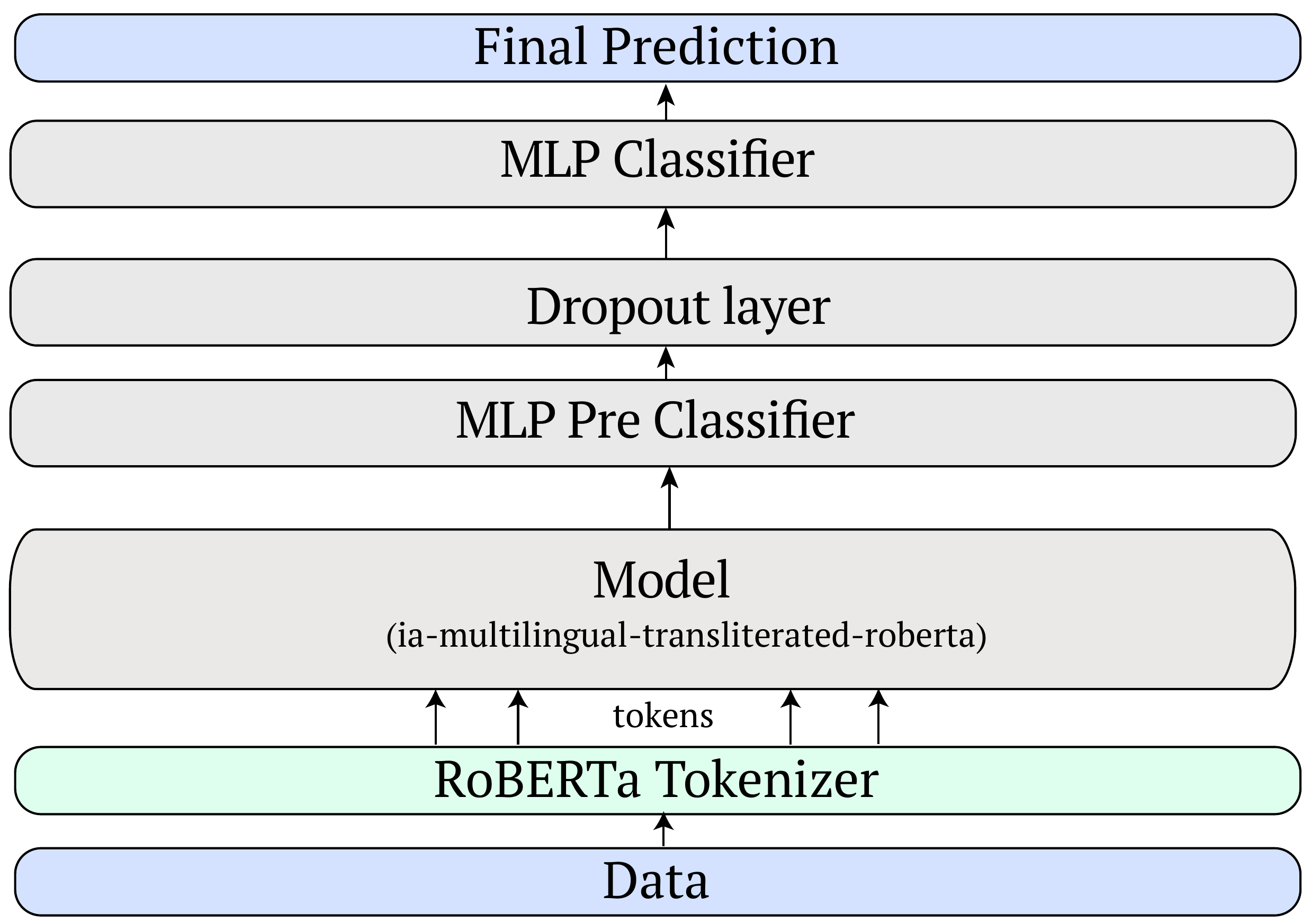}}
\caption{Model architecture, containing tokenizer, pre-trained model, classifier and other components}
\label{fig:WHAT}
\end{figure}

Dialectal and cultural variations, as well as close linguistic ties across languages, further complicate NLP for South Asian languages. This workshop will explore these challenges, focusing on issues around linguistic and cultural diversity, encoding and orthographic challenges, and the resource limitations that slow down technological advancement. By addressing these hurdles, we hope to move closer to improving NLP tools for South Asian languages, helping to preserve their linguistic and cultural heritage.

We decided to participate in Subtask B, Hate Speech Detection, which focuses on identifying the presence of hate speech in Devanagari-scripted languages, and Subtask C, Target Identification, which aims to pinpoint specific targets of hate speech, such as individuals, organizations, or communities.
These tasks align closely with broader challenges in South Asian language processing, where complex sociolinguistic contexts, dialect diversity, and multilingualism often complicate language understanding, especially in identifying and managing harmful content online. By advancing these areas, we contribute to addressing critical gaps in natural language understanding for South Asian languages.

\section{System Description}
\subsection{Problem and Dataset}
\textbf{Subtask B: Hate Speech Detection.}
Hate speech on online platforms can lead to social harm, reinforce biases, and promote hostility within communities \cite{parihar2021hate,wasi-2024-explainable}. For Devanagari-scripted languages such as Hindi \cite{jafri2024chunav,jafri2023uncovering}, Marathi \cite{kulkarni2021l3cubemahasent}, Nepali \cite{rauniyar2023multi, thapa2023nehate}, Bhojpuri \cite{ojha2019english}, and Sanskrit \cite{aralikatte2021itihasa}, hate speech detection is particularly challenging due to limited resources, diverse dialects, and complex linguistic structures. In this task, participants must develop a model to detect hate speech within text data written in Devanagari script. Given a dataset with binary annotations indicating the presence or absence of hate speech, the objective is to accurately classify each text instance to either "contains hate speech" or "does not contain hate speech." Success in this task will enhance the ability to monitor and mitigate harmful online content in Devanagari-scripted languages, fostering safer digital environments. The dataset includes 19,019 training tweets, 4,076 validation tweets, and 4,076 test tweets \cite{thapa2025nludevanagari,sarves2025chipsal}.

\noindent
\textbf{Subtask C: Target Identification for Hate Speech}
Understanding hate speech in Devanagari-scripted languages extends beyond detection to identifying specific targets, which can include individuals, organizations, or communities. Targeted hate speech poses unique threats by directly influencing public perception of vulnerable groups. In this task, participants are required to classify hate speech instances by identifying the intended target category—whether it is an "individual," "organization," or "community." This task aims to provide a deeper understanding of hate speech dynamics in South Asian languages, supporting more nuanced, targeted content moderation efforts across multilingual online platforms. The dataset for this task comprises 2,214 training tweets, 474 validation tweets, and 475 test tweets.

\subsection{Model Design and Development}

We designed a \texttt{MultilingualRobertaClass} model, which is a deep neural network architecture designed to address classification task in multilingual contexts, with a specific focus on Devanagari-scripted and transliterated languages. At its core, the model leverages the pretrained transformer model \texttt{ia-multilingual-transliterated-roberta}\footnote{https://huggingface.co/ibm/ia-multilingual-transliterated-roberta} \cite{dhamecha-etal-2021-role,liu2019robertarobustlyoptimizedbert} from IBM as its primary language representation layer. This layer, denoted by \texttt{self.l1}, captures contextualized embeddings of multilingual text, drawing on shared syntactic and semantic structures that span multiple languages. By leveraging a transformer model pretrained on multilingual data, the architecture is especially adept at handling languages with similar linguistic roots or transliterated forms, which are common in South Asian languages.

The model's classification head comprises a sequence of linear and non-linear transformations designed to adapt RoBERTa’s output \cite{dhamecha-etal-2021-role,liu2019robertarobustlyoptimizedbert} for binary classification. The output of the transformer’s CLS token, which serves as a global representation of the input text, is first passed through a fully connected layer, \texttt{self.pre\_classifier}, which maintains the 768-dimensional space. A ReLU activation function is then applied to introduce non-linearity, enhancing the model’s ability to capture complex language patterns. To reduce overfitting, a dropout layer with a 0.3 rate is included, which stochastically zeroes out 30\% of the neurons during training. This sequence of transformations allows the model to distill important semantic features from the multilingual embeddings while preserving relevant linguistic nuances for classification.

Finally, the classification layer, \texttt{self.classifier}, maps the 768-dimensional representation to a single scalar output, which undergoes a sigmoid activation to yield a probability score between 0 and 1. This score represents the model’s confidence in assigning the input to a specific binary class, suitable for tasks such as hate speech detection. By combining a pretrained multilingual transformer with a lightweight, fully connected classifier head, the \texttt{MultilingualRobertaClass} model balances generalizable language representation with efficient binary classification. This design is particularly beneficial in applications involving languages with high dialectal diversity and transliterated scripts, where robust language understanding and precise classification are paramount.

\begin{table*}[t]
\centering
\caption{Ablation Study Results for Subtask B}
\label{tab:stabB}
\scalebox{.95}{
\begin{tabular}{lccccc}
\hline
\textbf{Metric}                 & \textbf{Original} & \textbf{Sequence Length (128)} & \textbf{Learning Rate (\(10^{-5}\))} & \textbf{Batch Size (8)} \\ \hline
Accuracy                         & 0.8221            & 0.8050                        & 0.8150                                      & 0.8180 \\ \hline
Weighted Precision               & 0.7984            & 0.7800                        & 0.7900                                      & 0.7950\\ \hline
Weighted Recall                  & 0.8221            & 0.8100                        & 0.8180                                      & 0.8200\\ \hline
Weighted F1 Score                & 0.8097            & 0.7940                        & 0.8040                                      & 0.8070\\ \hline
Micro Precision                  & 0.8221            & 0.8050                        & 0.8150                                      & 0.8180\\ \hline
Micro Recall                     & 0.8221            & 0.8100                        & 0.8180                                      & 0.8200 \\ \hline
\end{tabular}}
\end{table*}

\begin{table*}[h!]
\centering
\caption{Ablation Study Results for Subtask C}
\label{tab:stabc}
\scalebox{.95}{
\begin{tabular}{lcccc}
\hline
\textbf{Metric}                 & \textbf{Original} & \textbf{Sequence Length (128)} & \textbf{Learning Rate (\(10^{-5}\))} & \textbf{Batch Size (8)} \\ \hline
Accuracy                         & 0.7321            & 0.7150                        & 0.7250                                      & 0.7300                  \\ \hline
Weighted Precision               & 0.7394            & 0.7250                        & 0.7350                                      & 0.7380                  \\ \hline
Weighted Recall                  & 0.7321            & 0.7200                        & 0.7300                                      & 0.7310                  \\ \hline
Weighted F1 Score                & 0.7326            & 0.7180                        & 0.7280                                      & 0.7300                  \\ \hline
Micro Precision                  & 0.7321            & 0.7150                        & 0.7250                                      & 0.7300                  \\ \hline
Micro Recall                     & 0.7321            & 0.7200                        & 0.7300                                      & 0.7310                  \\ \hline
\end{tabular}}
\end{table*}

\subsection{Implementation Details}
Our implementation begins by loading and preprocessing the dataset using \texttt{pandas} \cite{reback2020pandas} for data handling, and tokenizing input text with a transformer-compatible tokenizer from the \texttt{transformers} library \cite{wolf2020huggingfacestransformersstateoftheartnatural}, setting the maximum sequence length to 256 tokens. The data is split into training, validation, and test sets. For the model, a pre-trained transformer model (such as BERT \cite{devlin-etal-2019-bert}) from the \texttt{transformers} library is employed as the base, with a linear layer added for classification. Key hyperparameters include a learning rate of \(2 \times 10^{-5}\), batch sizes of 16 for training and 64 for evaluation, and a training duration of 2 to 5 epochs. The optimizer used is AdamW, chosen specifically for its suitability in fine-tuning transformers, alongside a linear learning rate scheduler with warm-up steps to adjust the learning rate dynamically during training.

The training process uses cross-entropy loss, suitable for classification, and evaluates performance with accuracy, precision, recall, and F1-score at the end of each epoch to monitor improvement. Additionally, gradient clipping is likely employed to stabilize the training process, although exact values are not specified. In the evaluation phase, the model generates predictions on validation and test sets, calculating final metrics to gauge model effectiveness. Predicted labels are derived from the logits generated by the model, and results, including performance metrics, are saved or displayed for analysis. This systematic approach ensures model performance is accurately tracked and evaluated throughout the training process.

\section{Experiments}
\subsection{Validation Set Results}
The validation results show that Subtask B consistently outperforms Subtask C across all metrics. Subtask B has higher accuracy (0.8221 vs 0.7321), precision (0.7984 vs 0.7394), recall (0.8221 vs 0.7321), and F1 score (0.8097 vs 0.7326). Both tasks have the same micro precision and recall values, indicating that Subtask B's model performs better in all aspects related to classification and error reduction. The substantial difference in scores suggests that Subtask B might be a simpler or more suitable task for the model, whereas Subtask C may involve more complexity or challenging data.

\subsection{Ablation Studies}
\textbf{Subtask B.} 
Table \ref{tab:stabB} presents the ablation study results for Subtask B. It shows that, reducing the sequence length to 128 tokens causes a slight drop in accuracy (from 0.8221 to 0.8050), suggesting that Subtask B benefits from longer sequences to capture more context. Lowering the learning rate to $10^{-5}$ also results in a small decrease in performance across all metrics, with accuracy dropping to 0.8150, indicating that the model's convergence is slower but still effective. Decreasing the batch size to 8 leads to a modest reduction in performance, with accuracy at 0.8180, implying that smaller batch sizes might affect the model's ability to estimate gradients effectively. Overall, the changes in hyperparameters have a moderate impact, with sequence length being the most influential, while the learning rate and batch size exhibit relatively smaller effects on Subtask B's performance.

\noindent
\textbf{Subtask C.}  
Table \ref{tab:stabc} presents the ablation study results for Subtask C. It shows that,. Reducing the sequence length to 128 tokens leads to a noticeable decrease in all metrics, particularly accuracy, which drops from 0.7321 to 0.7150. This suggests that Subtask C relies heavily on longer sequences to capture important context, and reducing the sequence length limits its ability to make accurate predictions. Lowering the learning rate to $10^{-5}$causes a slight decrease in performance across the board, with accuracy falling to 0.7250. This shows that a lower learning rate, while stabilizing training, may slow down convergence slightly. Reducing the batch size to 8 results in minimal changes in accuracy and other metrics, suggesting that Subtask C is relatively stable with smaller batch sizes. Overall, sequence length has the most significant impact on Subtask C's performance, with the learning rate and batch size having relatively smaller, but still noticeable, effects.

\begin{table}[t]
\centering
\caption{Test Results for Subtask B and Subtask C}
\label{tab:test-results}
\begin{tabular}{lcc}
\hline
\textbf{Metric}                 & \textbf{Subtask B} & \textbf{Subtask C} \\ \hline
Recall                          & 0.6547     & 0.5839     \\ \hline
Precision                       & 0.7106     & 0.5910     \\ \hline
F1 Score                           & 0.6762      & 0.5816    \\ \hline
Accuracy                        & 0.8840     & 0.6611    \\ \hline
\end{tabular}
\end{table}

\section{Evaluation}
The test results for Subtask B and Subtask C, presented in Table \ref{tab:test-results}, highlight significant differences in performance. In Subtask B, our model performs exceptionally well, achieving an accuracy of 0.8840, precision of 0.7106, recall of 0.6547, and an F1 score of 0.6762. These results indicate that the model is highly effective at detecting hate speech, with strong accuracy and a good balance between precision and recall, minimizing both false positives and false negatives. In contrast, Subtask C shows lower performance across all metrics, with an accuracy of 0.6611, precision of 0.5910, recall of 0.5839, and an F1 score of 0.5816. This suggests that identifying the specific targets of hate speech is a more challenging task, likely due to the increased complexity of categorizing targets like individuals, organizations, or communities. The lower results in Subtask C highlight the need for further model optimization or additional data to improve its performance in target identification.

\section{Discussion}
We believe, this research has a profound societal impact, fostering safer and more inclusive online spaces by advancing hate speech detection and target identification in South Asian languages. By addressing the challenges posed by the region's linguistic diversity and sociocultural nuances, it empowers platforms to better manage harmful content and protect vulnerable individuals and communities. Identifying specific targets of hate speech enables targeted interventions, offering support to marginalized groups while improving content moderation strategies. Beyond regional significance, the methodologies developed can inform global efforts to tackle online toxicity in multilingual contexts, promoting equitable digital environments, amplifying underrepresented voices, and contributing to social harmony and justice..

\section{Conclusion}
This paper presents a comprehensive approach to tackling two key subtasks in the domain of hate speech detection and target identification in Devanagari-scripted languages. We work, leveraging a pretrained multilingual transformer architecture to handle the complexities of hate speech detection in diverse South Asian languages. Our model demonstrated strong performance in Subtask B (Hate Speech Detection), with high accuracy and balanced precision and recall, effectively identifying hate speech instances. However, in Subtask C (Target Identification), the performance was lower, highlighting the additional challenges associated with identifying specific targets of hate speech. These results suggest that while hate speech detection can be effectively addressed with current approaches, more work is needed for robust target identification, which requires deeper understanding and handling of the nuanced sociolinguistic context in South Asian languages.

\section*{Limitations}
From an implementation perspective, one limitation of our work is the reliance on a pretrained multilingual transformer model, which may not fully capture the nuances of Devanagari-scripted languages and transliterated forms. Additionally, the model's architecture could benefit from further optimization, such as incorporating more specialized layers or attention mechanisms, to enhance its ability to handle complex linguistic structures and improve performance in Task C (Target Identification).

\section*{Acknowledgments}
We extend our heartfelt gratitude to the members and advisors of the Computational Intelligence and Operations Laboratory (CIOL) for their invaluable support throughout the preparation of this work and during the experimental phase. We also sincerely thank Wahid Faisal for his technical assistance with data augmentation, which greatly contributed to the success of this study.

\bibliography{our_work}




\end{document}